\documentclass[runningheads]{llncs}
\usepackage[T1]{fontenc}

\usepackage{graphicx}
\usepackage{multirow}
\usepackage{textpos}
\usepackage{listings}
\usepackage{xcolor}
\usepackage[utf8]{inputenc}
\usepackage{subfigure}
\usepackage{amsmath}
\usepackage{hyperref}
\usepackage{amssymb}
 \usepackage{pdfpages}

\makeatletter
\newcommand\notsoscript{\@setfontsize\notsoscript{8.5}{9.5}}
\makeatother

\begin{document}
\title{Reconstructing Historical Manuscripts through MSI: The Potential of Contrast in Assessing Image Quality and Legibility}

\titlerunning{The Potential of Contrast in Assessing Image Quality and Legibility}

\author{Anna Breger\inst{1}\orcidID{0000-0001-8878-5743} }
\authorrunning{A. Breger}

\institute{\vspace{-0.2cm}
$^1$Department of Applied Mathematics and Theoretical Physics, University of Cambridge, Cambridge, UK\\
}
\maketitle            
\vspace{-0.5cm}
\begin{abstract}
Digital restoration of historical manuscript images aims to improve readability while preserving the authenticity of cultural heritage documents. However, evaluating quality of restored manuscripts remains challenging, where readability is often subjective and expert annotations are scarce. This study investigates the suitability of contrast-based image quality measures to assess quality and legibility of reconstructed manuscript images from multi-spectral imaging. Two experiments were conducted with publicly-available data sets, facilitating manual quality scores by experts and full-reference image quality measures as reference evaluations. The results show that potential contrast achieves the highest correlation with expert ratings, while contrast-to-noise ratio demonstrates the strongest agreement with full-reference quality measures. Overall, contrast-based measures consistently outperform general image quality measures, demonstrating their potential as objective indicators of manuscript legibility and reconstruction quality.\let\thefootnote\relax\footnotetext{\notsoscript{Accepted at 17th IAPR International Workshop on Document Analysis Systems at 20th International Conference on Document Analysis and Recognition (ICDAR), Vienna 2026.}}

\end{abstract}
\vspace{-0.8cm}
\section{Introduction}
Digital restoration of historical manuscripts remains a central topic in document image analysis due to severe degradations that affect many cultural heritage collections. Common degradations such as ink fading, bleed-through and paper aging significantly reduce legibility and hinder both human interpretation and downstream processing tasks such as transcription and recognition. In recent years, substantial progress has been made with data-driven and deep learning based approaches for document analysis and enhancement \cite{dlhist} and through task-aware objectives have also improved readability of hand-written texts, see e.g.~\cite{KHAMEKHEMJEMNI2022108370}. Moreover, multi-spectral imaging (MSI) has emerged as a powerful non-invasive tool for the analysis and restoration of historical manuscripts by acquiring images across ultraviolet, visible, and near-infrared wavelengths. Together with advanced post-processing, MSI has demonstrated its power in revealing information that has otherwise not been visible, see e.g.~\cite{8810072},\cite{Janke2024SecondLook},\cite{Easton2010Standardized}. 

Automated evaluation of text legibility remains a challenging and largely open problem \cite{10.1007/978-3-030-86334-0_32}, although highly needed for task-driven restoration frameworks and larger amounts of data. Common image quality assessment (IQA) measures are primarily designed for natural images and often fail to capture legibility in degraded documents. In order to understand suitability of IQA measures, dedicated evaluation protocols and comprehensive data sets would be needed for existing as well as newly developed approaches. However, available data sets that enable the assessment of IQA methods for historical manuscript legibility remain very limited. Some data sets have been developed, but never published, e.g.~\cite{8480372}, others had been made available in the past but disappeared over time, e.g.~\cite{SHAHKOLAEI2018199}. The few available data sets include the subjective assessment framework SALAMI, that provides expert-based legibility annotations for manuscript images \cite{10.1007/978-3-030-68787-8_5}, as well as the Parchment data set, that includes artificially degraded parchment patches and their untreated references, in its original form \cite{10.1093/llc/fqv036} and modified \cite{Brenner2019ArtificiallyDegradedManuscripts}. Both data sets are based on MSI of the degraded manuscripts. 

Potential Contrast (PC) has been proposed in \cite{potcontr} as a task-dependent IQA measure that estimates the maximum achievable contrast between pixel classes under arbitrary intensity transformations. Unlike traditional contrast measures, it explicitly accounts for the possibility that relevant information may become visible only after suitable grayscale transformations, making it particularly relevant for cultural heritage applications by incorporating the possibility that the user might adjust the contrast in the viewing process. 
Its formulation allows user-guidance through annotated pixel sample sets and has been shown to be effective in identifying informative structures in degraded document images, e.g. \cite{10.1371/journal.pone.0178400}. Recently, it has been modified to be independent from the underlying data type, cf.~\cite{arxiv2505.01388}, and throughout our experiments we will employ this Normalized Potential Contrast (NPC). 

Despite its promising properties, the suitability of (N)PC as an IQA measure for historical manuscript legibility has not yet been systematically investigated. In particular, its relationship to text legibility and its behavior compared to other IQA measures when judging manuscript enhancement quality remains unclear. Given the increasing reliance on objective measures for evaluating restoration pipelines, we aim to provide first insights here. We design two experiments with the described data sets, analyzing the rank correlation of NPC values and expert ratings as well as reference-based quality evaluation across common degradations such as ink fading and bleed-through. For comparison we also report the contrast measure contrast-to-noise ratio (CNR), simple image statistics such as the root mean square contrast (RMSC) and entropy, as well as common no-reference (NR) IQA measures BRISQUE \cite{mittal2012brisque}, NIQE \cite{mittal2013niqe} and PIQE \cite{venkatanath2015piqe}. CNR can be understood as a direct complementing contrast measures by using the same sample masks in the computation as NPC. BRISQUE, NIQE and PIQE have been developed for natural images and do not require any reference information.  
\vspace{-0.2cm}
\section{Data and Methods} \label{data}
\subsection{Experiment 1: SALAMI Data \cite{Brenner2020SALAMI}\cite{10.1007/978-3-030-68787-8_5}}
\vspace{-0.2cm}
The SALAMI dataset (Subjective Assessments of Legibility in Ancient Manuscript Images), published in 2020, contains grayscale manuscript images ($900 \times 900$ pixels) derived from MSI data of 48 historical manuscripts, where 5 output reconstructions of 50 distinct image regions are shared publicly. The resulting 250 images had been annotated region-wise by 20 experts with philology and paleography background regarding legibility, yielding score maps for each image. The data set serves as a benchmark for developing and evaluating quantitative measures of legibility and digital restoration quality in historical manuscript imaging. A first IQA analysis with the data set was published by the data collectors in \cite{10.1007/978-3-030-86334-0_32}, reporting the Spearman Rank Correlation Coefficients (SRCC) of several quality measures and the mean score maps, including simple image statistics, text detection and NR IQA measures, over several window sizes and dedicated text areas. (N)PC had not been included in the study as it requires defined foreground and background pixels, which also holds true for CNR. 

For our complementing experiments we choose from the $50$ distinct images the ones that contain writing over the whole image and where at least one of the provided $5$ versions of each image allows automated text and background extraction via direct thresholding yielding objective binary annotation masks. This selection process results in $55$ test images ($11$ images with $5$ versions each), in which we compute $10 000$ random patches of the size $200 \times 400$ to account for the region-wise score maps, respectively. In Figure \ref{salaminpc} we show two examples of random patches and corresponding mean score maps. 
\vspace{-0.4cm}
\subsection{Experiment 2: Parchment Data \cite{Brenner2019ArtificiallyDegradedManuscripts} and Random Reconstructions}
\vspace{-0.3cm}
The Parchment data set, published in 2019, and available at \cite{brenner2019artificially}, is a modified subset of the original data set \cite{10.1093/llc/fqv036} published in 2017. The image data is based on multispectral acquisitions of 18th-century parchment manuscript fragments written with iron-gall ink, imaged before and after controlled artificial degradation treatments. It holds 22 parchment patches ($720 \times 720$) with different treatment, including untreated control samples, and provides registered multispectral image data with $21$ spectral bands from 400 to 950 nm. The degradation procedures simulate common forms of manuscript deterioration, enabling direct evaluation of the degraded images with their intact counterparts serving as a ground truth. 

Here, we chose $4$ patches (208R, 305R, 309R, 602V) of the provided data set, restricting the experiment to patches with severe degradations/illegible parts after treatment, as well as requiring to have some signal information of the degraded bits in at least one of the $21$ MSI bands, i.e.~enforcing possible reconstruction of the illegible notation, see examples in Figure \ref{parchex}. 
Furthermore, we cropped the images to the degraded parts, e.g.~if the lower half became illegible after the treatment, but the upper half remained perfectly intact, then only the lower half was evaluated to ensure feasible results representing the degraded regions. Binary masks for text and background were created manually on small regions therein, employing a heuristically chosen MSI band with sufficient remaining information, mimicking a real use case without available ground truth. For each image 15 minutes were set to create the masks in the software GIMP, imitating a research situation with constraint timelines. 

Next, we use the MSI data of the treated patches to create with random orthogonal projections a set of $100000$ reconstructed grayscale outputs per image to be evaluated for quality beyond the 25 provided outputs per image. \vspace{-0.6cm}
\subsubsection{Dimension reduction with orthogonal projections.}
\textit{Let $x \in \mathbb{R}^d$ be a high-dimensional image. A random lower-dimensional representation is given by $px \in V$, 
where $V\subset \mathbb{R}^d$ denotes a $k$-dimensional linear subspace with $k<d$ and $p \in \mathbb{R}^{d \times d}$ an orthogonal projection distributed according to the unique orthogonally invariant probability measure on the Grassmannian. The dimension $d$ is reduced to $k$ in practice via elements $q \in \mathbb{R}^{k \times d}$ of the Stiefel manifold with $q^Tq = p$.} 

Random projections can be computed efficiently via the QR composition, cf.~\cite{chikuse2003statistics}, and it has been shown that they give a good representation of the whole space whilst preserving important properties, cf.~\cite{ortho}. Here, we do have $d=21$ MSI bands and reduce the images directly to a grayscale output, i.e.~$k=1$, in which case it corresponds to the uniform distribution on the sphere. 

With ground truth images available, we employ $3$ full-reference (FR) IQA measures that have shown stable behavior to assess structural information or, in particular, text legibility, as references of quality by comparing the random reconstruction to the untreated ground truth data. Namely, we employ HaarPSI \cite{reisenhofer2018haarpsi} based on Haar wavelets, Pearson correlation \cite{Brenner2019ArtificiallyDegradedManuscripts} and the multi-scale SSIM \cite{wang2003multi}. As suggested in the original paper \cite{Brenner2019ArtificiallyDegradedManuscripts}, we implement the evaluation invariant to polarity. Then, the SRCC is computed between the tested quality measures and these $3$ reference measures. As an additional experiment, we determine for each of the four images the highest-ranked MSI band among the $21$ available bands for every tested quality measure. Please find the results in the supplementary material.
\vspace{-0.5cm}
\begin{figure}[htbp!]
\centering
\subfigure[npc = 0.56]{\includegraphics[width=0.35\textwidth]{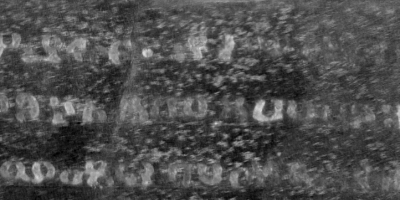} }
\subfigure[npc = 0.13]{ \includegraphics[width=0.35\textwidth]{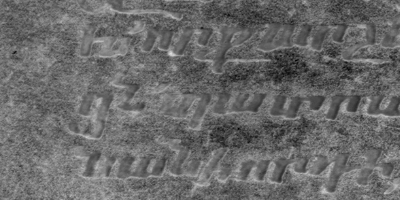}} \\
\subfigure[mean score map]
{\includegraphics[width=0.35\textwidth]{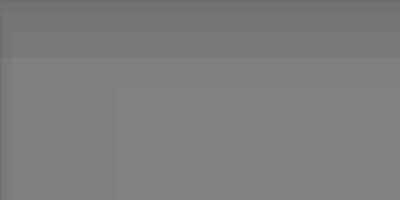}} \ 
\subfigure[mean score map]
{ \includegraphics[width=0.35\textwidth]{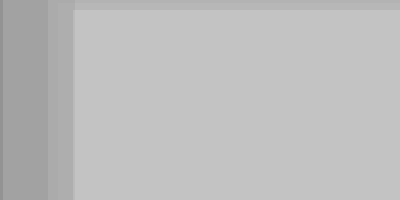}} 
\caption{Visual examples of two random patches (a)(b) in the SALAMI data set with corresponding NPC values and mean score maps (c)(d) stored as uint8 images, where white (255) denotes the best and black (0) the worst score. The values of NPC are in [0,1], with $1$ denoting the best judgement. The left example shows good matching between the NPC and the expert ratings with an average score over the map of $126.54$, the right example shows a pitfall of NPC, judging the quality poorly although it is rated with an average score of $188.95$ by the experts.
}
\label{salaminpc}
\end{figure}

\begin{figure}
\centering
\subfigure[cnr = 1.51]{ \includegraphics[width=0.3\textwidth]{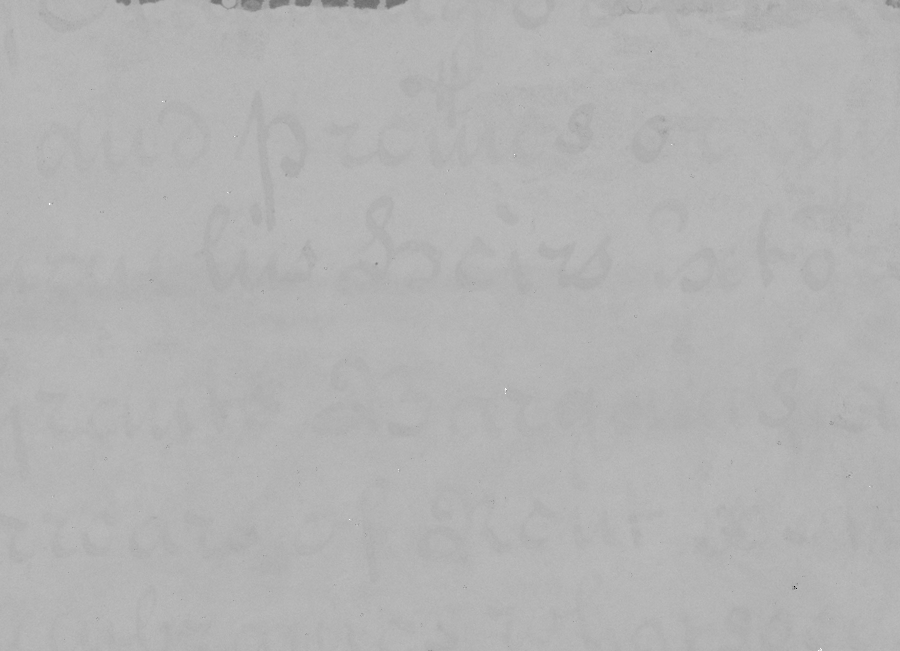}} 
\subfigure[cnr = 2.28]{\includegraphics[width=0.3\textwidth]{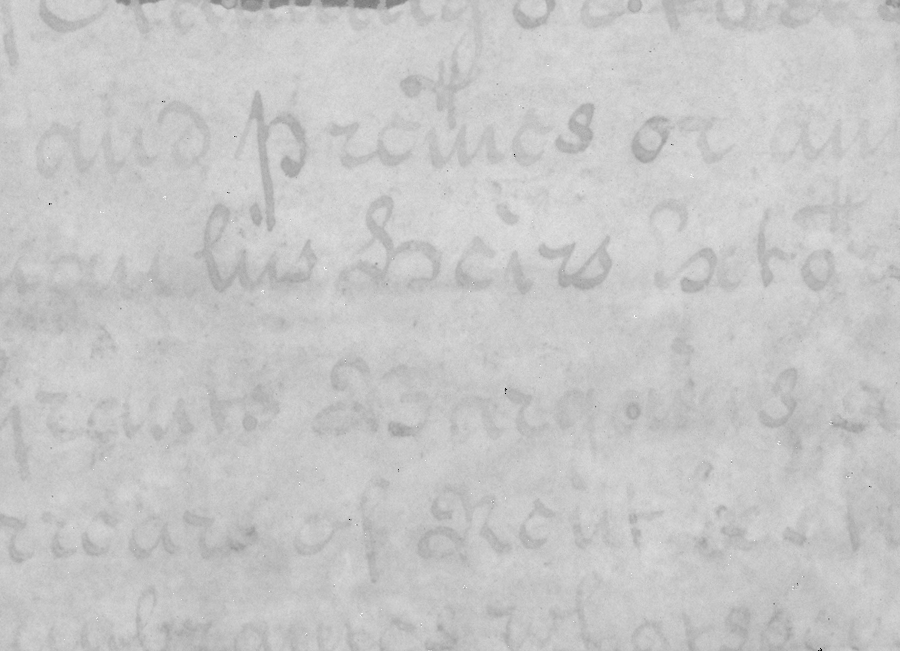}} 
\subfigure[cnr = 3.03]{ \includegraphics[width=0.3\textwidth]{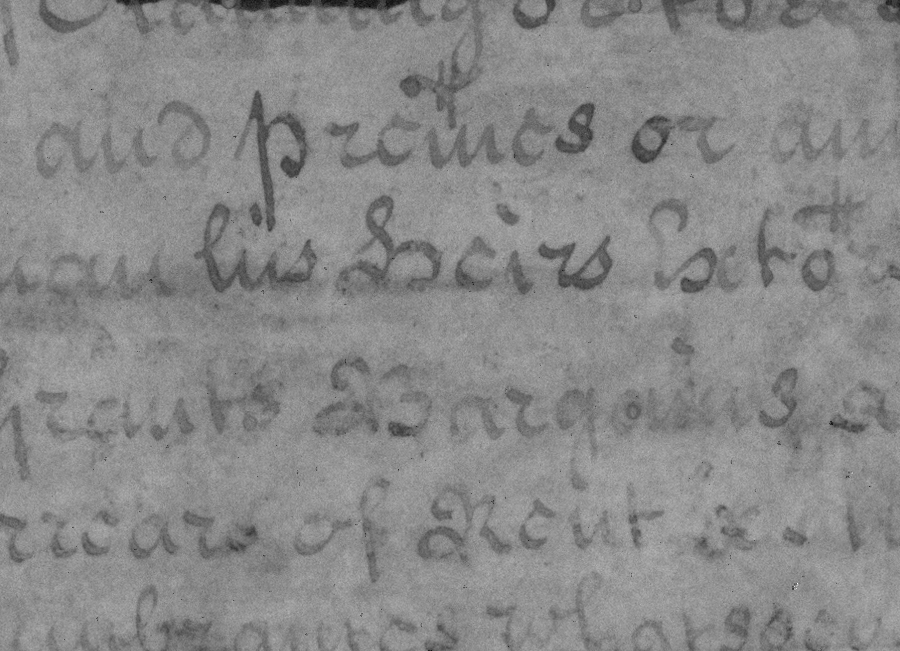}} \\
\subfigure[npc = 0.40]{ \includegraphics[width=0.3\textwidth]{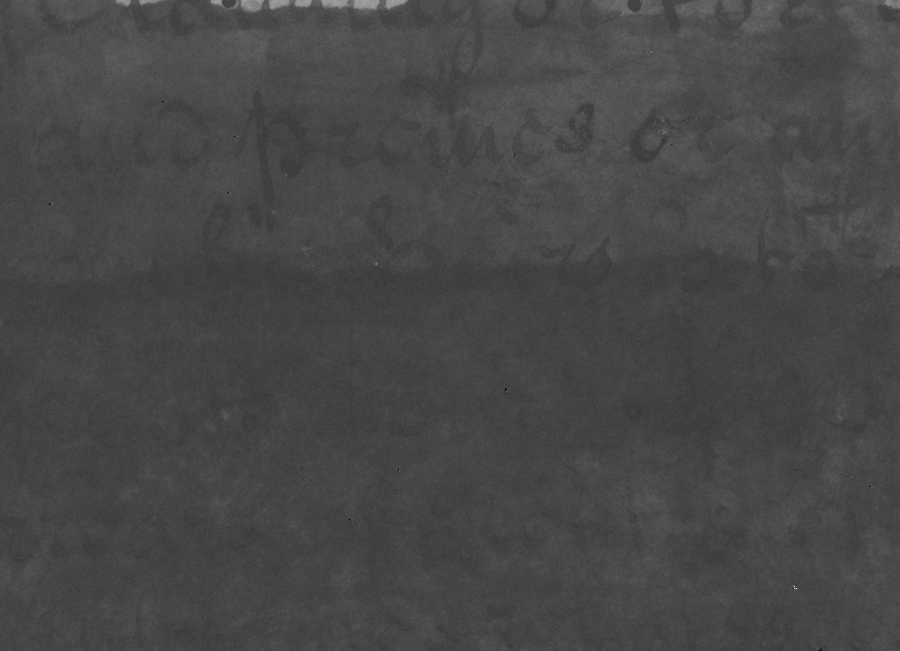}} 
\subfigure[npc = 0.59]{\includegraphics[width=0.3\textwidth]{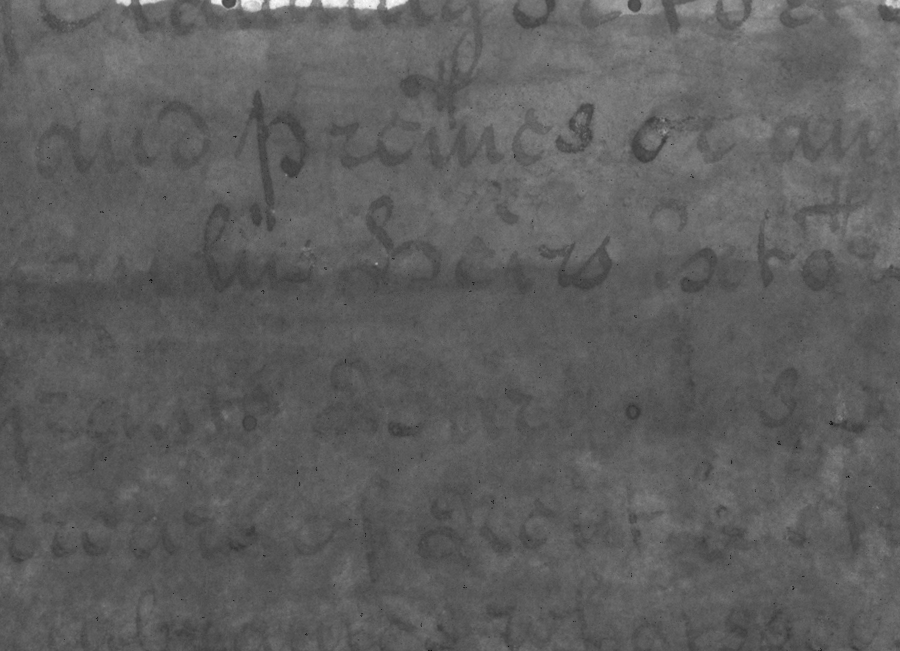}} 
\subfigure[npc = 0.77]{\includegraphics[width=0.3\textwidth]{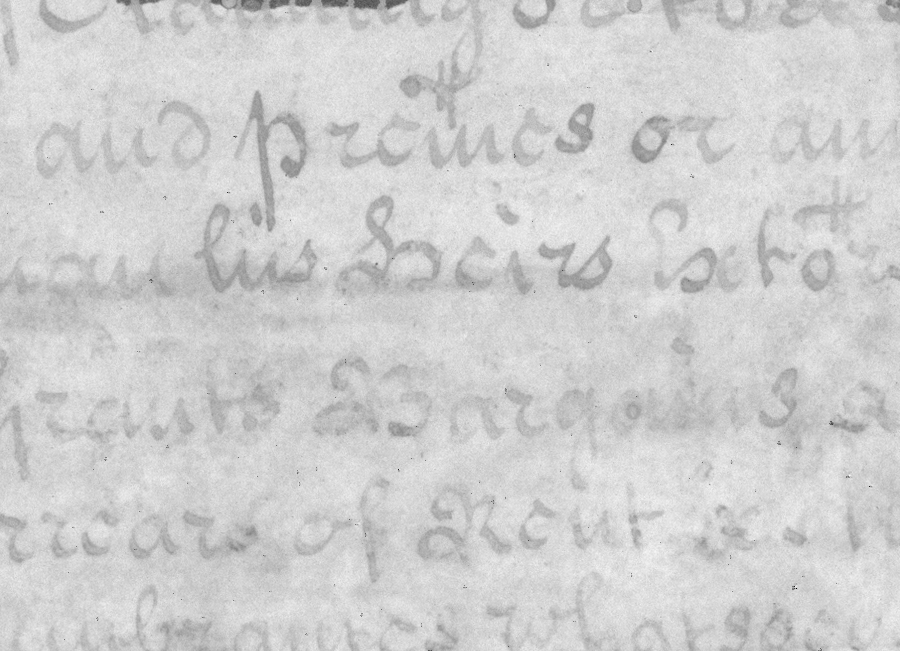}} 
\caption{Visual examples of random reconstructions of the MSI data of fol.305R in the Parchment data set. The images are ordered in each row in regards to the quality assessment by CNR (top) and NPC (bottom), with increasing quality values from the left to the right, where the right gives the highest quality value, respectively, within the $100 000$ random reconstructions. 
\vspace{-0.3cm}
}
\label{parchex}
\end{figure}
\section{Results}
\vspace{-0.2cm}
In the quantitative results we will state the absolute Spearman Rank Correlation Coefficient (SRCC) of the measures' quality evaluation and the references, indicating how well the tested measures align with the reference judgement. In the qualitative analysis we exemplify results of the highest achieving IQA measures.
\vspace{-0.4cm}
\subsection{Experiment 1}
In Table \ref{tab:metrics} the correlation results of the IQA measures and mean score maps are shown. The highest result was achieved by NPC with a mean correlation of $0.6784$ over the $10 000$ random patches and $0.6561$ for the full image comparison.

\begin{table}[b]
\vspace{-0.2cm}
\centering
\scalebox{1.0}{
\begin{tabular}{c||c|c|c|c|c|c|c}
IQA method & BRISQUE & CNR & ENTROPY & \ NPC \ \ & NIQE \qquad & PIQE & RMSC\\
\hline
\hline
SRCC \textit{mean$_{10k}$}\quad & 0.0562 \quad & 0.5332 \quad & 0.4805 \quad & \textbf{0.6806} \quad  & 0.2031 \quad & 0.1630 \quad & 0.5896 \quad \\
\scriptsize \textit{variance$_{10k}$}\quad & \scriptsize 0.0021 \quad & \scriptsize 0.0039 \quad & \scriptsize 0.0009 \quad & \scriptsize 0.0005 \quad & \scriptsize 0.0005 \quad & \scriptsize 0.0002 \quad & \scriptsize 0.0012 \quad \\
\hline
SRCC \textit{full}\quad & 0.1709 \quad & 0.4627 \quad & 0.5136 \quad & \textbf{0.6561} \quad & 0.2341 \quad & 0.1558 \quad & 0.6150 \quad \\
\end{tabular} }
\vspace{0.1cm}
\caption{Absolute SRCC values for the tested IQA measures and the mean annotation scores over all $55$ images with the $10000$ random patches of size $200 \times 400$ (mean and variance) as well as the full images. The highest correlation value is highlighted in bold.}
\label{tab:metrics}
\end{table}

In Figure \ref{salaminpc}, we visualize 2 random patches with their NPC score and the corresponding mean score maps. The first example shows a patch where the NPC value and score map align well, whereas in the second example NPC fails.

\subsection{Experiment 2}
\begin{figure}[htbp!]
\vspace{-0.5cm}
\centering
\subfigure[npc = 0.77]{\includegraphics[width=0.35\textwidth]{images/ExampleParchment/Example_npc7697_i15671.png}} 
\subfigure[npc = 0.77]{ \includegraphics[width=0.35\textwidth]{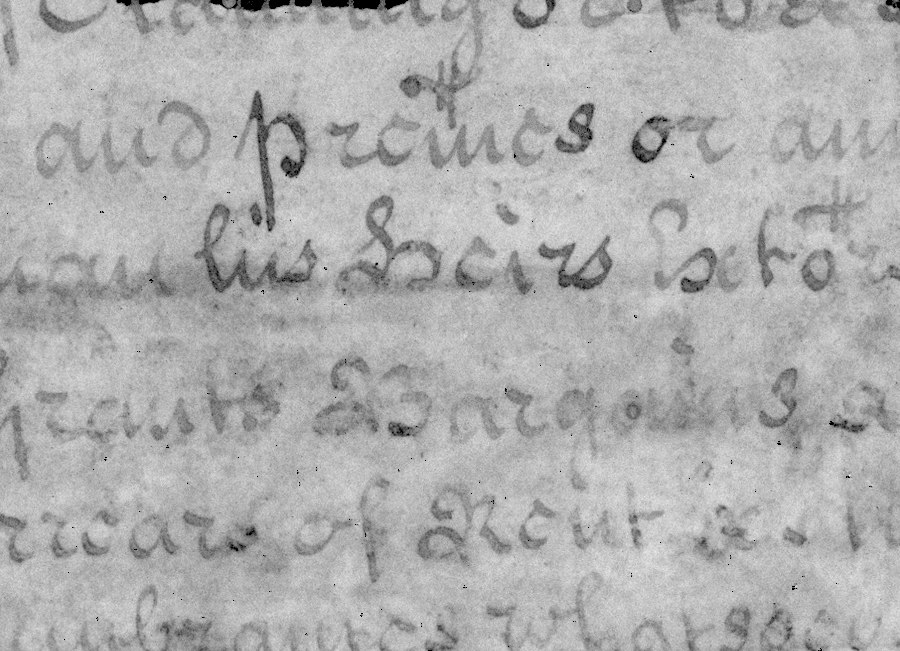}} 
\caption{NPC is invariant to linear brightness/contrast changes in the image. This property does not hold for the FR IQA measures it has been compared to in Experiment 2, influencing the correlation result.}
\label{npcexams}
\end{figure}

We report in Table \ref{tab:metrics2} the absolute SRCC over $100000$ random reconstructions for each of the $4$ images between the tested IQA measures and the chosen set of FR-IQA measures, i.e.  Haar-PSI, Pearson Correlation and MS-SSIM, serving as a reference judgement set. Moreover, in Figure \ref{parchex}, we show $3$ random reconstructions ordered by IQA values from CNR and NPC, the two measures that obtained the highest correlation results. It is important to note that NPC is invariant to linear brightness/contrast changes and we display in Figure \ref{npcexams} a visual result when adjusting the image with intensity transformations. 

\begin{table}
\centering
\scalebox{0.92}{
\begin{tabular}{c||c|c|c|c|c|c|c}
IQA & BRISQUE & CNR & ENTROPY & \ NPC \ \ & NIQE \qquad & PIQE & RMSC\\
\hline
\hline
Image 1 & .08,.10,.12 & \textbf{ .84,.83,.91} & .64,.55,.47 &  .76,.75,.86& .16,.27,.24 & .01,.23,.24 & .01,.03,.04 \\
Mean \quad & 0.10 & \textbf{0.86} & 0.55 & 0.79 & 0.22 & 0.16 & 0.03 \\
\hline
Image 2 & .13,.18,.18 & .84,\textbf{.95},\textbf{.94} & .90,.55,.75 &  \textbf{.85},.94,\textbf{.94}& .58,.32,.52 & .17,.01,.01 & .77,.34,.59 \\
Mean \quad & 0.16 & \textbf{0.91} & 0.73 & \textbf{0.91}  & 0.47 & 0.06 & 0.57 \\
\hline
Image 3 & .06,.06,.09 & \textbf{.86},\textbf{.96},\textbf{.88} & .66,.48,.64 &  .79,.76,.79& .68,.50,.67 & .42, .32, .43 & .55,.34,.53 \\
Mean \quad & 0.07 & \textbf{0.90} & 0.59 & .78 & 0.61 & 0.39 & 0.48 \\
\hline
Image 4 & .11,.09,.00 & \textbf{.53,.94,.58} & .23,.53,.07 &  .50,.82,.50& .50,.47,.49 & .29,.40,.33 & .22,.49,.06 \\
Mean \quad & 0.07 & \textbf{.68} & 0.27 & 0.61 & 0.49 & 0.34 & 0.26 \\
\end{tabular} }
\vspace{0.1cm}
\caption{Absolute SRCC values of the tested IQA measures and the reference measures HaarPSI, Pearson Correlation and MS-SSIM, respectively as well as the mean, over all $100 000$ random reconstructions. The highest correlation values are highlighted in bold.}
\label{tab:metrics2}
\end{table}
\vspace{-0.5cm}
\vspace{-0.8cm}
\section{Discussion and Limitations}
For Experiment 1 we can observe in Table \ref{tab:metrics} that NPC clearly yields the highest correlation to the mean score maps derived from the experts' ratings, for the random patches as well as the full images, with the patch-wise computation yielding a higher result. This behavior is expected, since the mean score maps were created region-wise and the NPC computation over the whole image might not reflect local quality changes. The low variance confirms stability across the random patches. It is notable that the 3 contrast measures (NPC, RMSC and CNR) yield the 3 best results, indicating that indeed contrast notions can be helpful to identify image restoration quality as well as legibility. Nevertheless, NPC by no means acts perfectly, which is reflected in the overall correlation value of 0.68 and possible mismatches as displayed in Figure \ref{salaminpc}.

For Experiment 2, which has a more complex design since manual annotations are not available in the employed data set, we can observe in Table \ref{tab:metrics2} that CNR outperforms all tested measures, followed by NPC. This different behavior in comparison to Experiment 1 is not surprising because NPC is invariant to intensity transformations, which aligns well with quality ratings where humans might adjust the intensity when viewing, but does not align necessarily well with FR IQA measures that rely on computation in the intensity scaling that is given. In Figure \ref{npcexams} we show the visual difference that can be obtained while NPC remains the same. This property can be useful in practice when it is possible to adjust the brightness/contrast of an image output, e.g.~with an image viewer. Figure \ref{parchex} demonstrates that both, CNR and NPC, produce meaningful orders of image quality. 

Limitations of this study include that expert-annotated quality scores were not available for the Parchment dataset; therefore, $3$ FR-IQA measures were used as references instead. While these measures capture meaningful aspects of image quality, they do not necessarily represent human expert judgment. In addition, NPC and CNR depend on manually created masks, introducing subjectivity into the evaluation. Finally, larger and more diverse datasets are needed to improve robustness and generalizability and to capture a wider range of manuscript degradations. Further research shall study the behavior of recent deep-learning based and other document-specific image quality measures and include different image transformations to better assess their suitability for legibility evaluation in historical manuscripts. Note that Experiment 2 adds to the study \cite{10.1007/978-3-030-86334-0_32}, where results with complementing image quality measures may be found.
\vspace{-0.3cm}
\section{Summary}
In conclusion, the results indicate that contrast-based measures are valuable tools for assessing image restoration quality and legibility in historical manuscripts, tested here on images derived from MSI data. NPC showed the strongest performance when compared to human-derived quality maps, while CNR achieved the best results against full-reference image quality measures. Although both measures demonstrate great potential for legibility evaluation, limitations in the study are given. Future research shall explore additional quality measures and larger datasets to improve the reliability and generalizability of the study.
\bigskip 

\noindent \textbf{Acknowledgments.} A.B. acknowledges funding by the Cambridge Centre for Data-Driven Discovery and Accelerate Programme (grant LEAH/011) through a donation by Schmidt Sciences. 

\noindent\textbf{Disclosure of Interests.} The author has no competing interests to declare. 

\tiny
\bibliographystyle{splncs04}
\bibliography{bibnew}
\vspace{-0.5cm}



\section*{Supplementary material}

# Supplementary Material to *Reconstructing Historical Manuscripts through MSI: The Potential of Contrast in Assessing Image Quality and Legibility*

Anna Breger[1][0000-0001-8878-5743]

[1]Department of Applied Mathematics and Theoretical Physics, University of Cambridge, Cambridge, UK

## A. 1. Formulas

**Spearman Rank Correlation** To assess the alignment of the manual ratings and the IQA measures' evaluations, we employ the Spearman Rank Correlation (SRCC), see e.g. [5]. In contrast to the Pearson Correlation Coefficient, which computes the linear relationship of two variables, the SRCC computes how strongly the ranks of the entries of a vector $x \in \mathbb{R}^n$ (here, containing the manual quality ratings by the expert graders) and $y \in \mathbb{R}^n$ (here, containing the image evaluation values provided by an IQA measure), correlate. The SRCC can be written as

$$\mathrm{SRCC}(x, y) = 1 - \frac{6\sum_{i=1}^{n}(\mathrm{R}(x_i) - \mathrm{R}(y_i))^2}{n(n^2-1)},$$

where $\mathrm{R}(x_i)$ denotes the rank of the $i$-th entry of the vector $x$, corresponding here to the rating of the $i$-th image.

**Contrast-to-noise ratio** The contrast-to-noise ratio (CNR) for an image $I \subset \mathbb{R}^{m\times n}$ and two image regions $A \subset I$ and $B \subset I$ can be written as

$$\mathrm{CNR} = \frac{|\mu_A - \mu_B|}{\sigma},$$

where $\mu_A$ denotes the mean signal intensity in the region $A$, analogously for $B$, and $\sigma$ the background noise. Here, $A$ corresponds to the foreground text, $B$ to the background and we employ the standard deviation of the background to compute $\sigma$, i.e. the intensity variation of the parchment/paper background. A high CNR then indicates a high separability between the writing and the manuscript substrate, and a low CNR indicates that text and background might be similar in intensity or that the manuscript substrate obscures the writing.

**(Normalized) Potential Contrast** Due to its complexity we refer to the original papers for the definition of potential contrast (PC), cf. [7], and normalized potential contrast (NPC), cf. [6]. Note that we employ in all experiments the same binary text and background masks for the computation of NPC and CNR.

## A.2. Additional Experiments with Parchment Data [3][1]

In the main paper, 100000 random reconstructions of the parchment MSI data were used to create a large test set with varying levels of reconstruction quality. Here, complimenting that experiment, we provide more detailed insights by analyzing the quality evaluation of the provided MSI data for each of the 4 images. In particular, for the chosen quality measures we compute the quality values of all 21 bands and compare which bands has been chosen as the best in regards to each quality measure. Again, the full-reference IQA measures HaarPSI, Pearson Correlation and MS-SSIM serve as the point of reference. In Table 1 the results are summarized and chosen bands are displayed in Figure 1. We can see that CNR and NPC were able to reliably pick the best band as chosen by the FR-IQA measures, and on the other hand, that the common IQA measures BRISQUE, NIQE and PIQE failed for that task. In the visualization we can observe that the choices of those measures, demonstrated with BRISQUE, are inconsistent and sometimes even choose a band with hardly any remaining signal information.

| IQA | FR-IQA | BRISQUE | CNR | ENTROPY | NPC | NIQE | PIQE | RMSC |
|---|---|---|---|---|---|---|---|---|
| Image 1 | (15, 15, 15) | 1 | **15** | 11 | **15** | 1 | 1 | 8 |
| Image 2 | (19, 19, 19) | 5 | **19** | 17 | 18 | 1 | 1 | 17 |
| Image 3 | (18, 17, 18) | 14 | **17** | 11 | **17** | 1 | 1 | 14 |
| Image 4 | (3, 19, 3) | 1 | 19 | 11 | **3** | 1 | 1 | 10 |

**Table 1.** MSI bands that were rated best by the tested quality measures and the FR-IQA measures HaarPSI, Pearson Correlation and MS-SSIM over the 21 MSI bands per image. Closest match (majority vote) to the FR-IQA choice is highlighted in bold. Note that bands have been ordered according to their scanning wavelengths, i.e. closer band numbering indicates a more similar acquisition wavelength.

Moreover, we tested if the untreated ground truth image would be correctly identified as best image when added to the MSI stack, see Table 2. For all tested quality measures CNR and NPC were the only measures that correctly identified the ground truth as the best quality for all 4 images. Note that the ground truth data was not used to create the binary annotation masks for CNR and NPC. FR-IQA measures have been excluded, since the reference for their computation is the ground truth image.

| IQA | BRISQUE | CNR | ENTROPY | NPC | NIQE | PIQE | RMSC |
|---|---|---|---|---|---|---|---|
| Image 1 | X | ✓ | X | ✓ | X | X | X |
| Image 2 | X | ✓ | ✓ | ✓ | ✓ | X | X |
| Image 3 | X | ✓ | ✓ | ✓ | ✓ | X | X |
| Image 4 | X | ✓ | X | ✓ | X | X | X |

**Table 2.** Measures are marked if they were able to identify the ground truth as the best quality image when added to the 21 MSI bands.

The results in the additional experiments support the trend in the experiments of the main paper, showing that NPC and CNR produce meaningful results when applied as quality measures for legibility, whereas common measures such as BRISQUE fail to provide consistent results for that task.

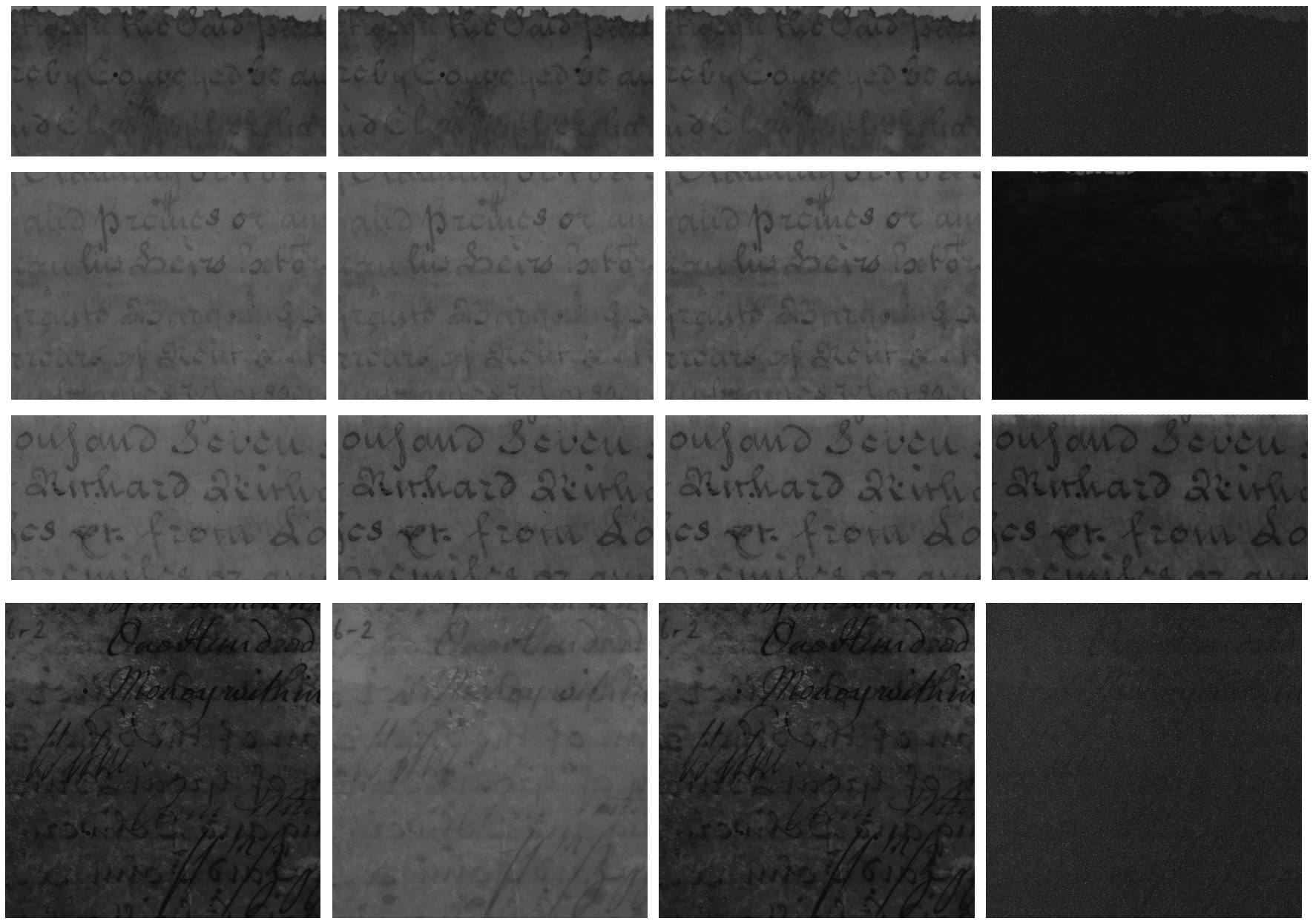

(a) FR-IQA (b) CNR (c) NPC (d) BRISQUE

**Fig. 1.** Visualization of results in Table 1 with best band picked by majority vote FR-IQA references (left), CNR (2nd col), NPC (3rd col) and BRISQUE (right). Images are cropped to the degraded parts as used in the analysis.

## A.3. Additional Figures

### SALAMI Data [2][4]

In Figure 2 we display 2 full example images with corresponding mean score maps from the expert annotations. Examples of random patches (size $200 \times 400$) are displayed in the main paper (Figure 1). In Figure 3 we show the binary masks derived by thresholding for the computation of CNR and NPC. Note that the patch-wise computation resembles a more realistic experimental set-up since text annotations for the whole image are usually too time-consuming.

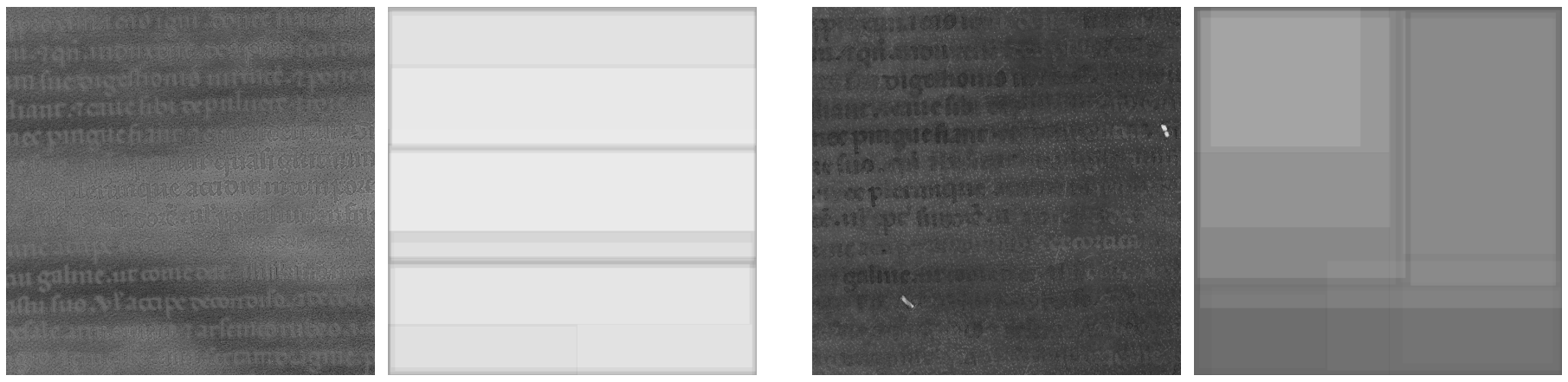

**Fig. 2.** Example images of the SALAMI data set with their corresponding provided mean score maps.

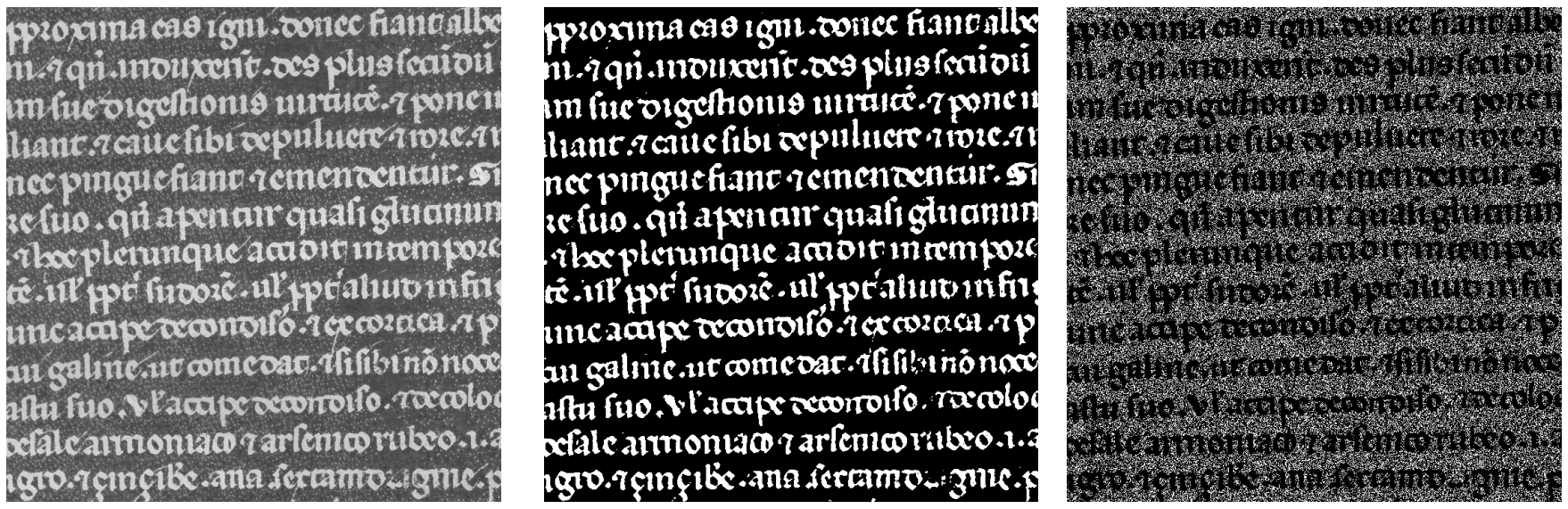

**Fig. 3.** Example reference image (left) of the SALAMI data set and derived binary masks for text (middle) and random background (right).

### Parchment Data

In Figure 5 we display the untreated ground truth images next to the treated (i.e. degraded) references and the manual annotations used for the computation of CNR and NPC. In Figure 4 we extend Figure 2 in the main paper, adding PIQE and BRISQUE to the visual evaluation, where we can observe that the provided ordering fails to represent visual quality and legibility.

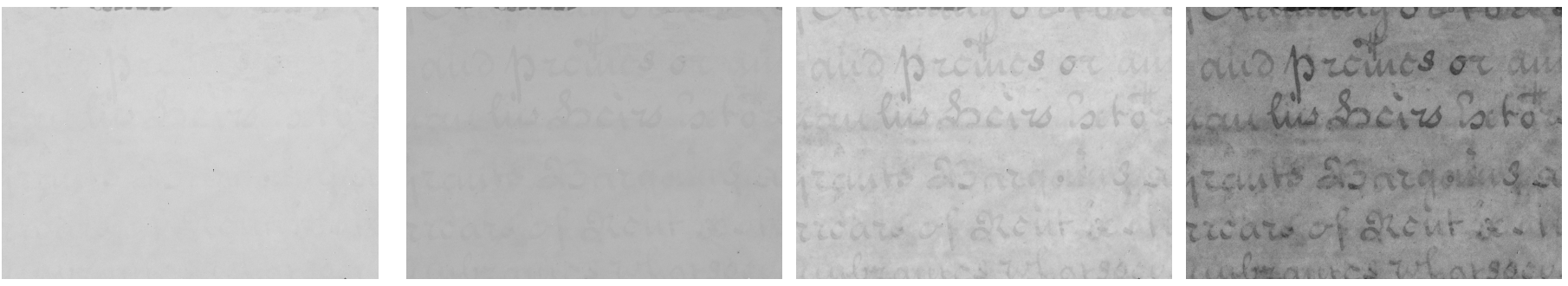

(a) cnr = 0.77 (b) cnr = 1.51 (c) cnr = 2.28 (d) cnr = 3.03

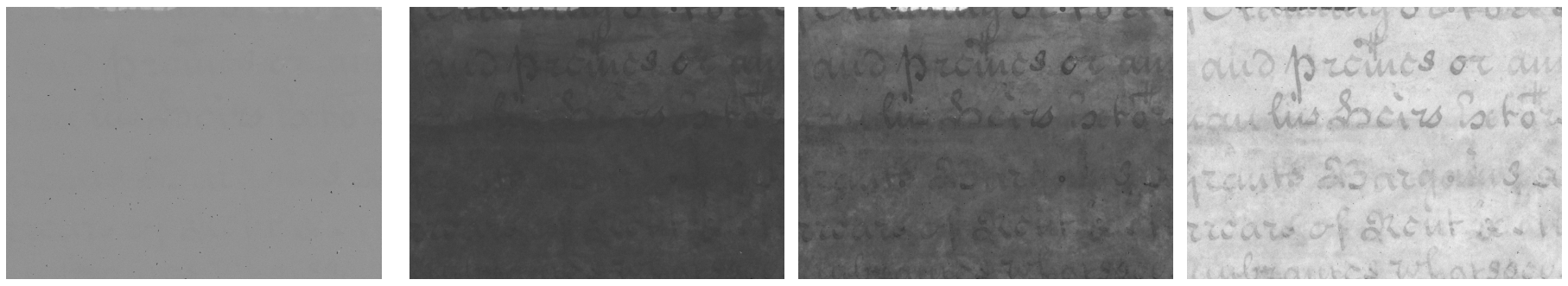

(e) npc = 0.20 (f) npc = 0.40 (g) npc = 0.59 (h) npc = 0.77

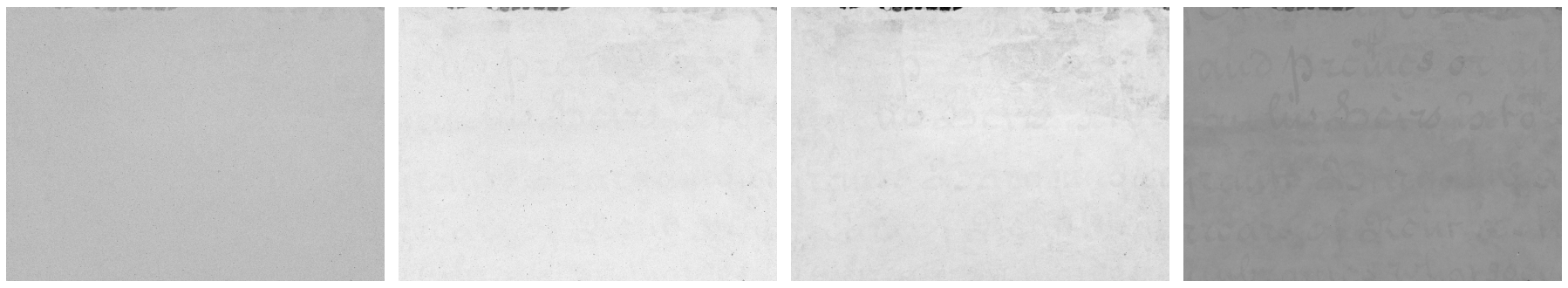

(i) piqe = 43.45 (j) piqe = 30.31 (k) piqe = 17.17 (l) piqe = 4.03

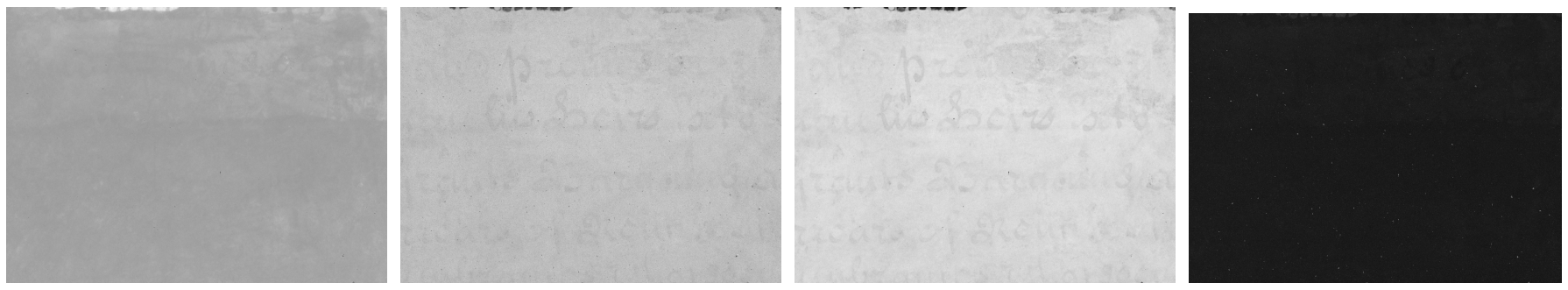

(m) brisque = 32.67 (n) brisque = 25.30 (o) brisque = 17.92 (p) brisque = 10.53

**Fig. 4.** Visual examples of random reconstructions of the MSI data of fol.305R. The images are ordered in each row in regards to the quality assessment by CNR (top), NPC (2nd row), PIQE (3rd row) and BRISQUE (bottom), with increasing quality values from the left to the right, where the right gives the highest quality value, respectively, within the 100000 random reconstructions. Note that it has been taken into account that PIQE and BRISQUE state quality in decreasing order.

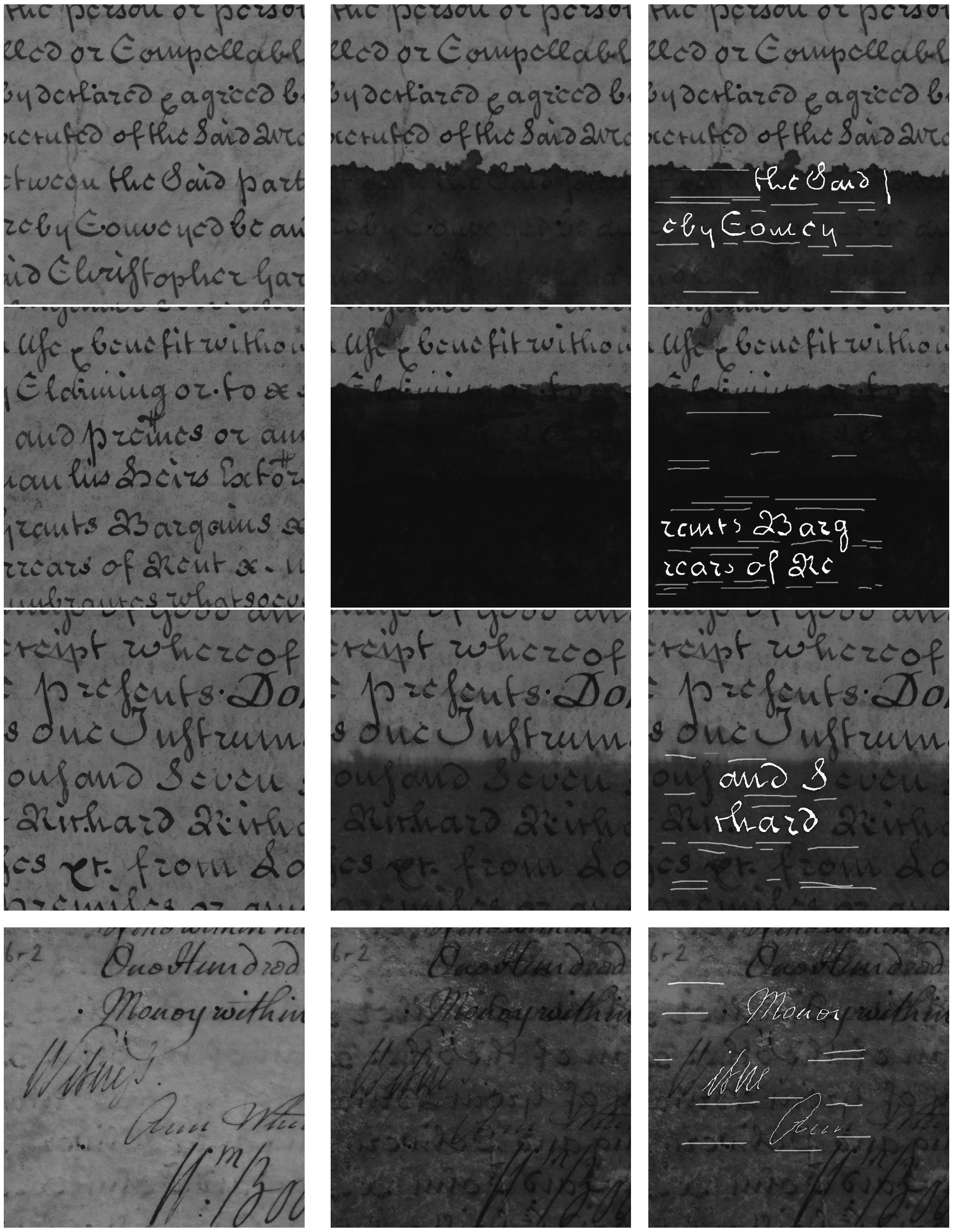

(a) Ground Truth (b) Treated Reference (c) Manual Annotation

**Fig. 5.** Ground truth (untreated) images, treated references and manual annotations of the 4 employed images in the Parchment data set: fol.208r, fol. 305r, fol.309r, fol.602v. Note that only the degraded parts were used for the analysis, see Figure 1 for the cropped versions of the images. The manual annotations are visualized with text in white and background in gray.


\end{document}